\documentclass[journal]{IEEEtran}
\def\@IEEEclspkgerror{\ClassError{IEEEtran}}
\usepackage{enumitem}
\usepackage{cite}
\usepackage{graphicx}
\usepackage{algorithm}
\usepackage{algorithmic}
\usepackage{amsmath}
\usepackage{amssymb}
\usepackage{amsthm}

\usepackage{soul}
\usepackage[dvipsnames]{xcolor}
\usepackage{hyperref}
\usepackage[dvipsnames]{xcolor}

\makeatletter
\newcounter{parenttheorem}

\makeatother

\def\BibTeX{{\rm B\kern-.05em{\sc i\kern-.025em b}\kern-.08em
    T\kern-.1667em\lower.7ex\hbox{E}\kern-.125emX}}
\usepackage{lineno}
\begin{document}

\title{A Hybrid Adaptive Velocity Aided Navigation Filter with Application to INS/DVL Fusion}
\author{Barak Or and Itzik Klein

\thanks{Barak Or and Itzik Klein are with the Hatter Department of Marine Technologies, Charney School of Marine Science, University of Haifa, Haifa, 3498838, Israel (e-mail: barakorr@gmail.com, kitzik@univ@haifa.ac.il).}}

\markboth{Preprint: A Hybrid Adaptive Velocity Aided Navigation Filter with Application to INS/DVL Fusion / Or and Klein}%
{Or and Klein: A Hybrid Adaptive Velocity Aided Navigation Filter with Application tto INS/DVL Fusion}
\maketitle

\begin{abstract}
Autonomous underwater vehicles (AUV) are commonly used in many underwater applications. Usually, inertial sensors and Doppler velocity log readings are used in a nonlinear filter to estimate the AUV navigation solution. The process noise covariance matrix is tuned according to the inertial sensors' characteristics. This matrix greatly influences filter accuracy, robustness, and performance. A common practice is to assume that this matrix is fixed during the AUV operation. However, it varies over time as the amount of uncertainty is unknown. Therefore, adaptive tuning of this matrix can lead to a significant improvement in the filter performance. In this work, we propose a learning-based adaptive velocity-aided navigation filter. To that end, handcrafted features are generated and used to tune the momentary system noise covariance matrix. Once the process noise covariance is learned, it is fed into the model-based navigation filter. Simulation results show the benefits of our approach compared to other adaptive approaches.  
\end{abstract}


\section{Introduction}\label{sec:introduction}
\IEEEPARstart{A}{utonomous} underwater vehicles (AUV) are commonly used in many underwater applications. Usually, an inertial navigation system (INS) and a Doppler velocity log (DVL) are used to estimate the AUV navigation state\cite{karaboga2013adaptive,tal2017inertial,elhaki2020robust,eliav2018ins}. Such fusion is commonly done using the error state implementation of the extended Kalman filter (es-EKF). In this filter, the process noise covariance is based on the characteristics of the inertial readings while the external aiding device (DVL) specification determines the measurement noise covariance. Those two covariances have a major influence on the filter accuracy, robustness, and performance \cite{bar2004estimation,farrell2008aided}.\\
A common practice is to assume constant covariances during the AUV operation. However, as the sensor characteristics are subject to change during operation and, in addition, physical constraints do not allow capturing the AUV dynamic optimally, those covariances should vary over time as the amount of uncertainty is unknown. Therefore, adaptive tuning of the process or measurement noise covariances can significantly improve the INS/DVL fusion performance. To cope with this problem, several adaptive model-based approaches were designed  \cite{lyu2021adaptive,kang2016adaptive,davari2016asynchronous,jwo2007adaptive,jaradat2014enhanced,tong2017adaptive,he2021adaptive}. The most common among them is based on the filter innovation process \cite{mehra1970identification}, where a measure of the new information is calculated at every iteration, leading to an update of the process noise covariance matrix. Still, the question of the optimal approach to tune the system noise covariance matrix is considered open and is addressed in detail in \cite{zhang2020identification}. \\
Recently, learning approaches were integrated into AUV navigation algorithms to improve their performance \cite{klein2022}. In \cite{cohen2022}, an end-to-end network was suggested to improve the accuracy of the DVL velocity estimation instead of using the model-based approach. Sea experiments results show an improvement of more than $50\%$. To cope with situations of missing beams, a DNN to regress the missing beams was suggested in \cite{yona2021compensating,nadav2022}. Later, a DNN model was used to cope with DVL failure scenarios and predict the DVL output \cite{li2021underwater}. In \cite{Yinging2020}, performance of INS/DVL fusion using a tightly coupled approach in situations of partial DVL beam measurement, was improved using a learning virtual beam aided solutions. Focusing on learning within the estimation process, a self-learning square root-cubature Kalman filter was proposed \cite{shen2020seamless}. This hybrid navigation filter enhanced the navigation accuracy by providing measurements, even in global navigation satellite system (GNSS) denied environments, by learning transfer rules of  the internal signals in the filter. In \cite{or2022learning}, recurrent neural networks were employed to learn the vehicle’s geometrical and kinematic features to regress the process noise covariance in a linear Kalman filter framework. In our previous work \cite{or2022hybrid}, we derived a hybrid learning framework for INS/GNSS fusion. Field experiments using a quadrotor showed an improvement of 25\% in the quadrotor position accuracy relative to model-based approaches.

In this paper, we propose a hybrid learning adaptive velocity-aided navigation filter. We rely on the model-based es-EKF and design machine learning (ML) models to tune the momentary system noise covariance matrix. To that end, handcrafted features (HCF) are fed into the learning model assisting it in finding the optimal parameters during the training procedure. Once the process noise covariance is learned, it is fed into the well-established, model-based Kalman filter.  To validate our proposed approach, AUV trajectories were created in simulation to create a dataset containing sensor readings and ground-truth values. Based on this dataset, we adopt a supervised learning (SL) approach using ensemble bagged trees \cite{gashler2008decision,myles2004introduction} to regress the process noise covariance. The advantages of the proposed approach lie in its hybrid fusion of the well celebrated es-EKF and learning approaches. By using the latter, we leverage their ability to generalize intrinsic properties appearing in sequential datasets and, therefore, better cope with varying conditions affecting the process noise values of the INS/DVL navigation filter.

The rest of the paper is organized as follows: Section II presents our proposed hybrid adaptive es-EKF with HCF. Section IV presents the results and Section V gives the conclusions of this study.
\section{A Hybrid Adaptive INS/DVL Navigation Filter}
\label{sec:Hybrid}
\begin{figure*}[ht]
\centering
{\includegraphics[width=0.8\textwidth]{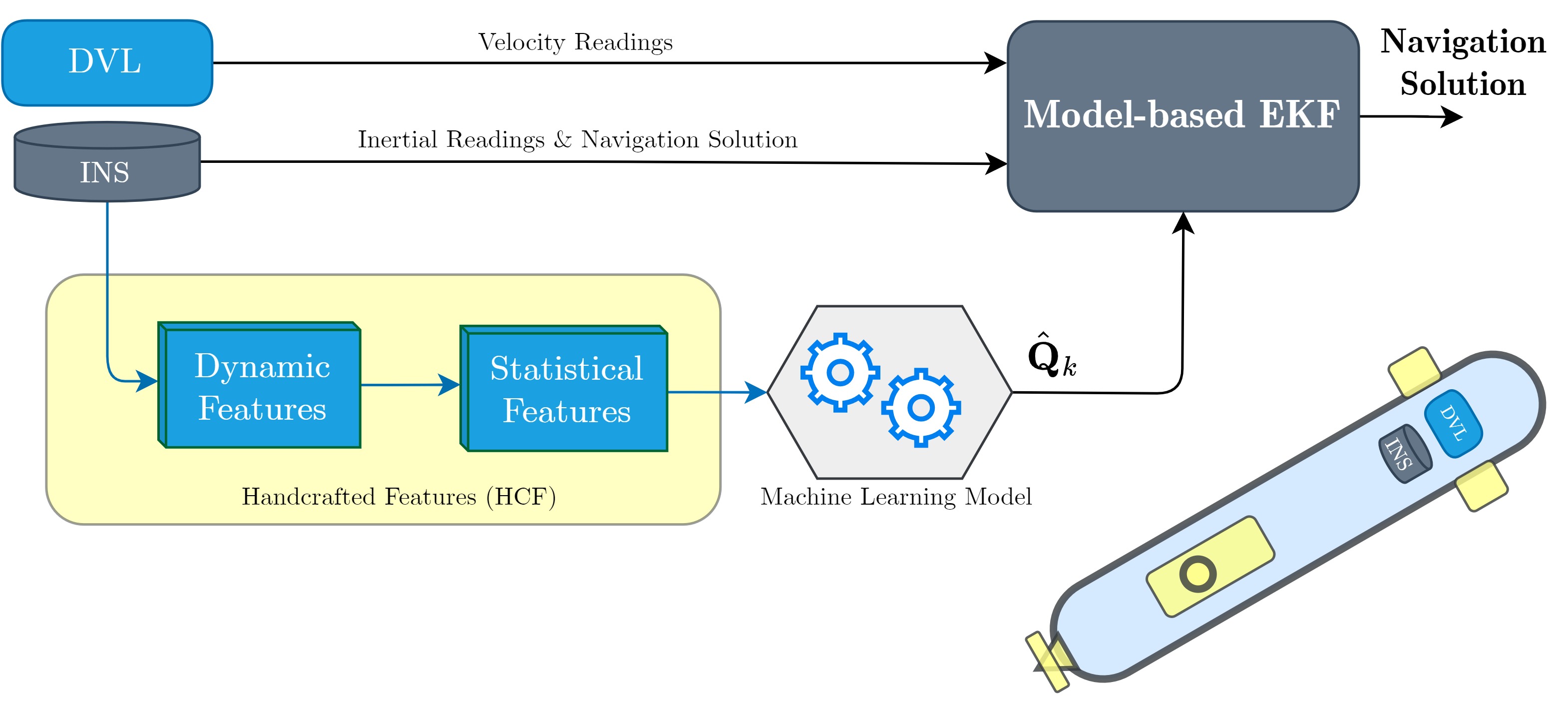}}
\caption{Boosting the hybrid adaptive navigation filter in the INS/DVL fusion with handcrafted features for online tuning of the process noise covariance matrix.}
\label{[Fig1a}
\end{figure*}

Relying on our hybrid learning adaptive navigation filter  \cite{or2022hybrid}, we employ HCF (instead of DNNs) as a tool to improve the accuracy of the process noise estimation in an INS/DVL fusion process and as a consequence improve the navigation performance. To that end, a classical ensemble bagged trees model is trained with statistical and dynamic features. Our proposed framework is presented in Fig.~\ref{[Fig1a}. 
\subsection{Data-Driven Based Adaptive Noise Covariance}
\label{sec:DataDriven}
It is assumed that the continuous process noise covariance matrix at time $k$, ${\bf Q}_k^c$, is a diagonal matrix with the following structure: 
\begin{equation}
\begin{array}{l}
{\bf{Q}}_k^c = \\
diag{\left\{ {q_f^{x*},q_f^{y*},q_f^{z*},q_\omega ^{x*},q_\omega ^{y*},q_\omega ^{z*},\varepsilon{{\bf{I}}_{1 \times 6}}} \right\}_k} \in{\mathbb{R}}{^{12 \times 12}},
\end{array}
\end{equation}
where the variance for each of the accelerometer axes is given by ${q_f^{i*}}$, for the gyroscope by ${q_\omega ^{i*}}$, and  $ i \in x,y,z$. The biases are modeled by random walk processes, and their variances are set to $\varepsilon=0.001$. Our goal is to find $q_k^*$ (for all axes) such that the speed error is minimized. To determine the process noise matrix, the problem is formulated as an SL task. Formally, we search for a model to relate an instance space, ${\cal X}$, and a label space, ${\cal Y}$. We assume that there exists a target function, $\cal F$, such that ${\cal Y} = {\cal F}\left( {\cal X} \right)$. \\
To that end, a general one-dimensional series of length $N$ of the inertial sensor (accelerometer or gyroscope) readings in a single axis is defined by 
\begin{equation}
{{\cal S}_k} = \left\{ {{{\bf{s}}_i}} \right\}_{i = k - N}^{k -1}.
\end{equation}
Thus, the SL task for adaptive determination of the process noise covariance means to find ${\cal F}$, given a finite set of labeled examples (inertial sensor readings), and corresponding process noise variance values: 
\begin{equation}\label{eq:sets}
\left\{ {{{\cal S}_k},q_k^*} \right\}_{k = 1}^M.
\end{equation}
where $M$ is the number of examples. The SL approach aims to find a function ${\tilde {\cal F}}$ that best estimates ${\cal F}$. To that end, for the training process, a loss function is defined to quantify the quality of ${\tilde {\cal F}}$ with respect to ${\cal F}$. The loss function is given by
\begin{equation}\label{eq:l1}
{\cal L}\left( {{\cal Y},\hat {\cal Y}\left( {\cal X} \right)} \right) \buildrel \Delta \over = \frac{1}{M}\sum\limits_{m = 1}^M {l{{\left( {y,\hat y} \right)}_m}},
\end{equation}
where $m$ is the example index. Minimizing $\cal L$ in a training/test procedure leads to the target function. 
In the problem at hand, the loss function in  \eqref{eq:l1} is given by
\begin{equation}
{l_m} \buildrel \Delta \over = {\left( {q_m^* - {\hat q}_m} \right)}^2,
\end{equation}
where ${\hat q}_m$ is the estimated term obtained by the learning model during the training process.
\subsection{Dataset Generation}
\label{sec:dataset}
\begin{figure}[ht]
\centering
{\includegraphics[width=0.4\textwidth]{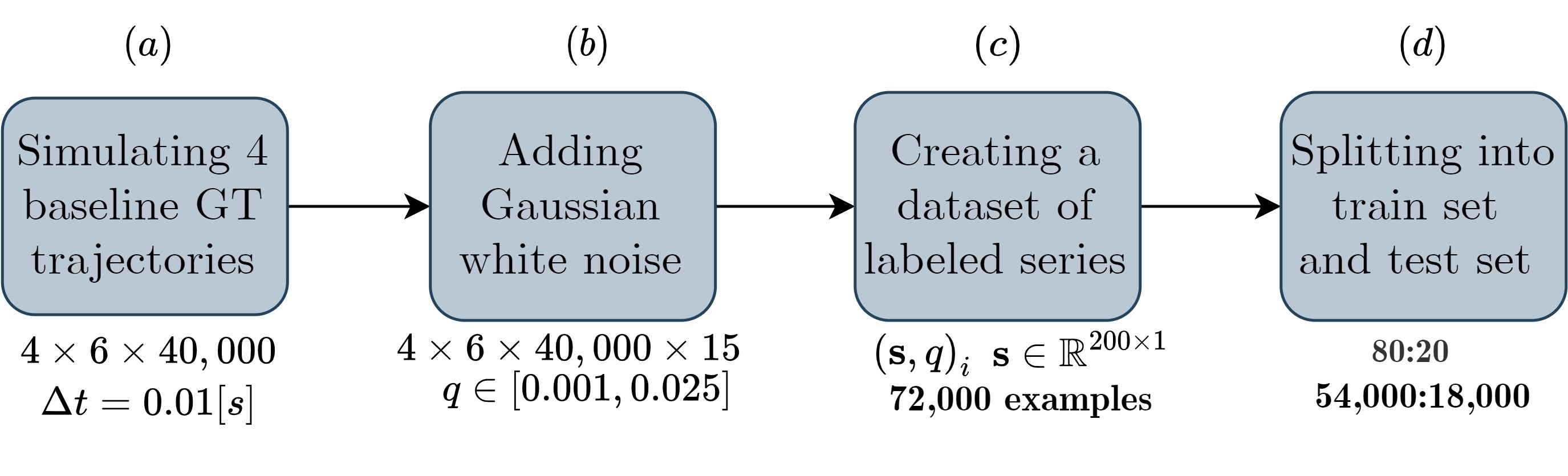}}
\caption{Dataset generation and pre-processing phase.}
\label{[Fig1}
\end{figure}

\begin{figure}[ht]
\centering
{\includegraphics[width=0.35\textwidth]{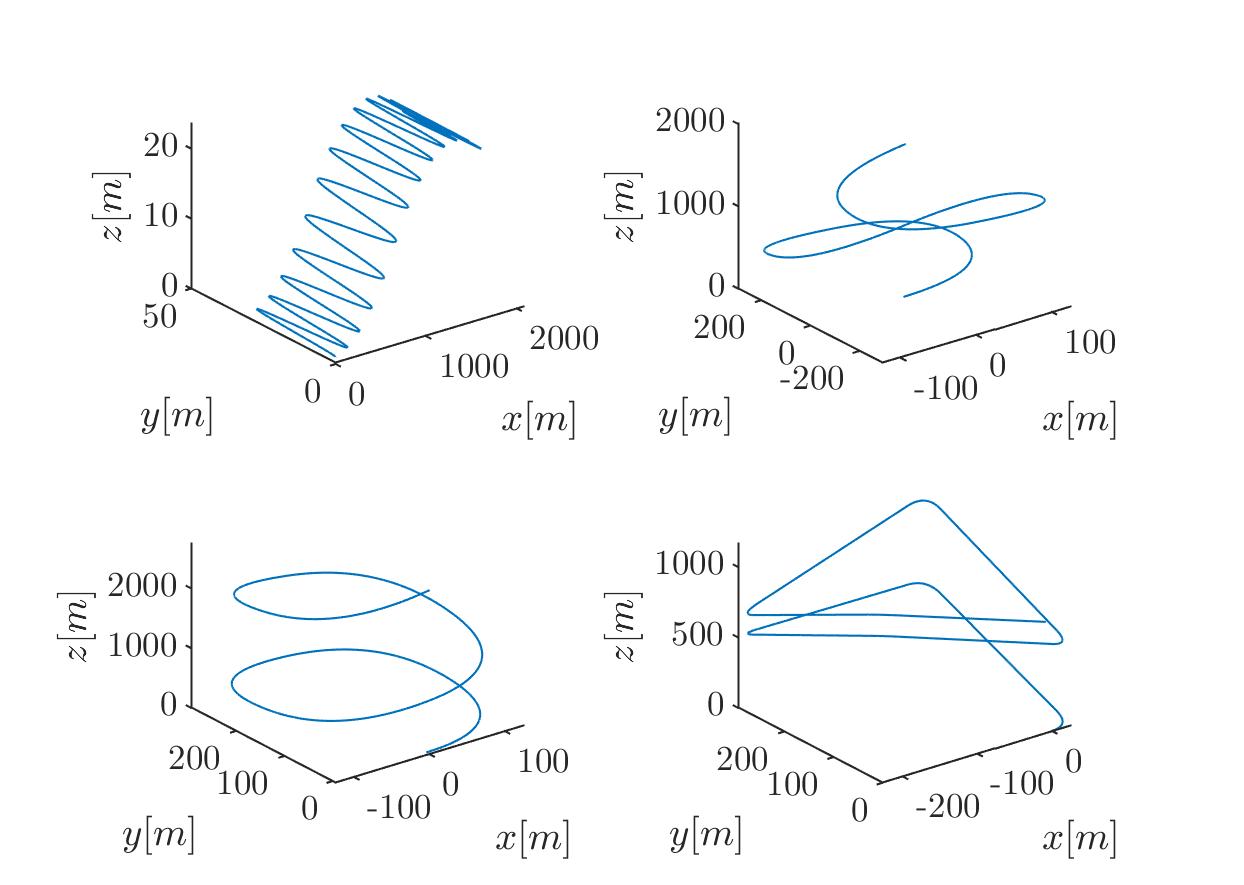}}
\caption{Our four baseline trajectories used to create the dataset. }
\label{fourbaseline}
\end{figure}

A stimulative-based dataset was generated leveraging our simulation designed in \cite{or2022hybrid}. To generate the dataset four different baseline trajectories were simulated and divided into train and test sets as illustrated in Figure~\ref{[Fig1}. The richness of the baseline trajectories, presented in  Figure~\ref{fourbaseline} allows the establishment of a model which is able to cope with unseen trajectories. Each baseline trajectory was created by generating ideal inertial readings for a period of $400$s in a sampling rate of $100$Hz, resulting in a sequence of $6\times40,000$ samples for each baseline trajectory.
To simulate real-world varying conditions, we added to each of the six inertial channels an additive zero-mean Gaussian noise with variance in the range of $q\in[0.001,0.05]$ and with 15 different values inside this interval.
Hence, each baseline trajectory has $15$ series of $6\times40,000$ noisy inertial samples. The justification to include a simple noise model lies in the momentary IMU measurement noise covariance sequence as a short time window is considered and thus allows characterizing the noise with its variance only. Next, the series length $N$ is chosen, and a corresponding labeled database is generated. Lastly, we choose $N=200$ samples and create batches corresponding to two seconds each with a total of $6\times200\times 15$ per baseline trajectory. Then, these batches are randomly divided into train and test sets in such a manner that all baseline trajectories are included both in the train and the test sets with a ratio of 80:20. In total, there are 72,000 examples; 57,600 in the train set and 14,400 in the test set. 
\subsection{Handcrafted features}
\label{sec:Handcrafted}
\begin{figure*}[h!]
\centering
{\includegraphics[width=0.9\textwidth]{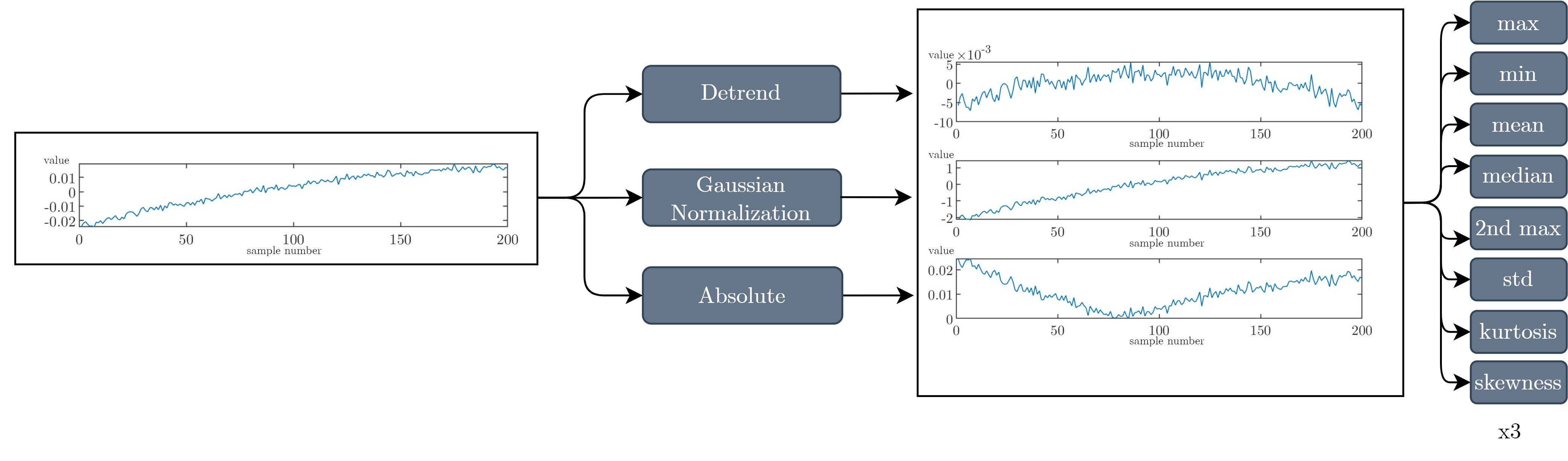}}
\caption{HCF Pipeline: readings of an inertial sensor are inserted into three high-level features and then, for each one of them, eight low-level features are calculated.}
\label{Fig61}
\end{figure*}

We propose a two-step approach for creating the HCF. At first, readings of an inertial sensor are plugged into three high-level features. Secondly, eight low-level features are calculated on each of the three top-level features resulting in 24 features for each inertial sensor, as shown in Figure~\ref{Fig61}.
The three top-level features are:
\begin{enumerate}
\item Detrend:
\begin{equation}
{f_1} = detrend\left( {\cal S} \right)
\end{equation}
A detrend calculates the trend of the sequence and removes it from the sequence. Then, it keeps only the differences in values from the trend. It allows cyclical and other patterns to be identified easier.
\item Gaussian normalization:
\begin{equation}
{f_2} = \left\{ {\frac{{{{\cal S}_j} - mean\left( {\cal S} \right)}}{{Std\left( {\cal S} \right)}}} \right\}_{j = 1}^{200}
\end{equation} 
\item Absolute values series:
\begin{equation}
{f_3} = \left\{ {\left| {{\cal S}_j^{}} \right|} \right\}_{j = 1}^{200}
\end{equation}
\end{enumerate}
The output of each of those three top-level features is a vector with 200 values. On that output, low-level features are calculated. Those include min, max, median, std, mean, kurtosis, skewness, and second max. An ensemble bagged trees model was trained with the HCF. It combines 30 decision trees with a minimum of 8 leaves each.
\subsection{Hybrid Learning Approach} \label{sec:HybridScheme}
Applying the suggested adaptive tuning approach in an online setting involves integrating the INS/DVL with the regressor as presented in Fig. \ref{[Fig1a}. Algorithm 1 describes the INS/DVL with process noise covariance learning in a real-time setup. The IMU signals are inserted both into the INS/DVL filter and the ML model. 
\begin{algorithm}[h!]
 \caption{Hybrid adaptive filter applied to INS/DVL}
 \begin{algorithmic}[1]
 \renewcommand{\algorithmicrequire}{\textbf{Input:}}
 \renewcommand{\algorithmicensure}{\textbf{Output:}}
 \REQUIRE ${\bf \omega}_{ib},{\bf f}_b,{\bf{v}}_{Aiding},\Delta t_0, \Delta \tau,T,tuningRate$
 \ENSURE  ${\bf{v}}^n,{\bf \varepsilon}^n$
 \\ \textit{Initialization} : ${\bf{v}}^n_0,{\bf{\varepsilon}}^n_0$
 \\ \textit{LOOP Process}
   \FOR {$t = 0$ to $T$}
   \STATE obtain ${\bf \omega}_{ib},{\bf f}_b$ (IMU readings)
  \STATE solve navigation equations (3)
  \IF {$(mod (t,\Delta \tau)$=0)}
  \STATE obtain ${\bf{v}}_{DVL}$ 
 \STATE  update navigation state using the es-EKF (8)-(14)
  \ENDIF
  \STATE Calculate HCF from IMU readings
  \STATE Calculate model using HCF and predict ${\bf Q}_{k+1}^c$ 
  \IF {$mod(t,tuningRate)=0$}
\STATE ${\bf{Q}}_{k + 1}^c \leftarrow {\bf{\hat Q}}_{k + 1}^c\left( {{{\cal S}_k}} \right)$
\ENDIF
  \ENDFOR
 \end{algorithmic}
 \end{algorithm}
\section{Analysis and Results} \label{sec:Results}
Two metrics are employed to evaluate the speed accuracy of the proposed approach:
\begin{itemize}
    \item Speed root mean squared error ($2^{nd}$ norm) for all three axes:
\begin{equation}
SRMSE = \sqrt {\frac{1}{y}\sum\limits_{k = 1}^{y}  {\sum\limits_{j \in \left\{ {N,E,D} \right\}}^{} {\delta {{\hat v}_{jk}}^2} } } .
\end{equation}
    \item Speed mean absolute error for all three axis:
\begin{equation}
SMAE = \frac{1}{y}\sum\limits_{k = 1}^{y} {\sum\limits_{j \in \left\{ {N,E,D} \right\}}^{} {\left| {\delta {{\hat v}_{jk}}} \right|}},
\end{equation}
\end{itemize}
where $y$ is the number of samples, $k$ is a running index for the samples, $j$ is the running index for the velocity component in the North, East, and Down (NED) directions, and $\delta {\hat v}_{jk}$ is the velocity error term.

For comparison to the proposed HCF-Ensemble model, we implement our deep-learning model, as described in \cite{or2022hybrid}, and refer to it as DNN. 

\begin{figure}[h!]
\centering
    {\includegraphics[width=0.4\textwidth]{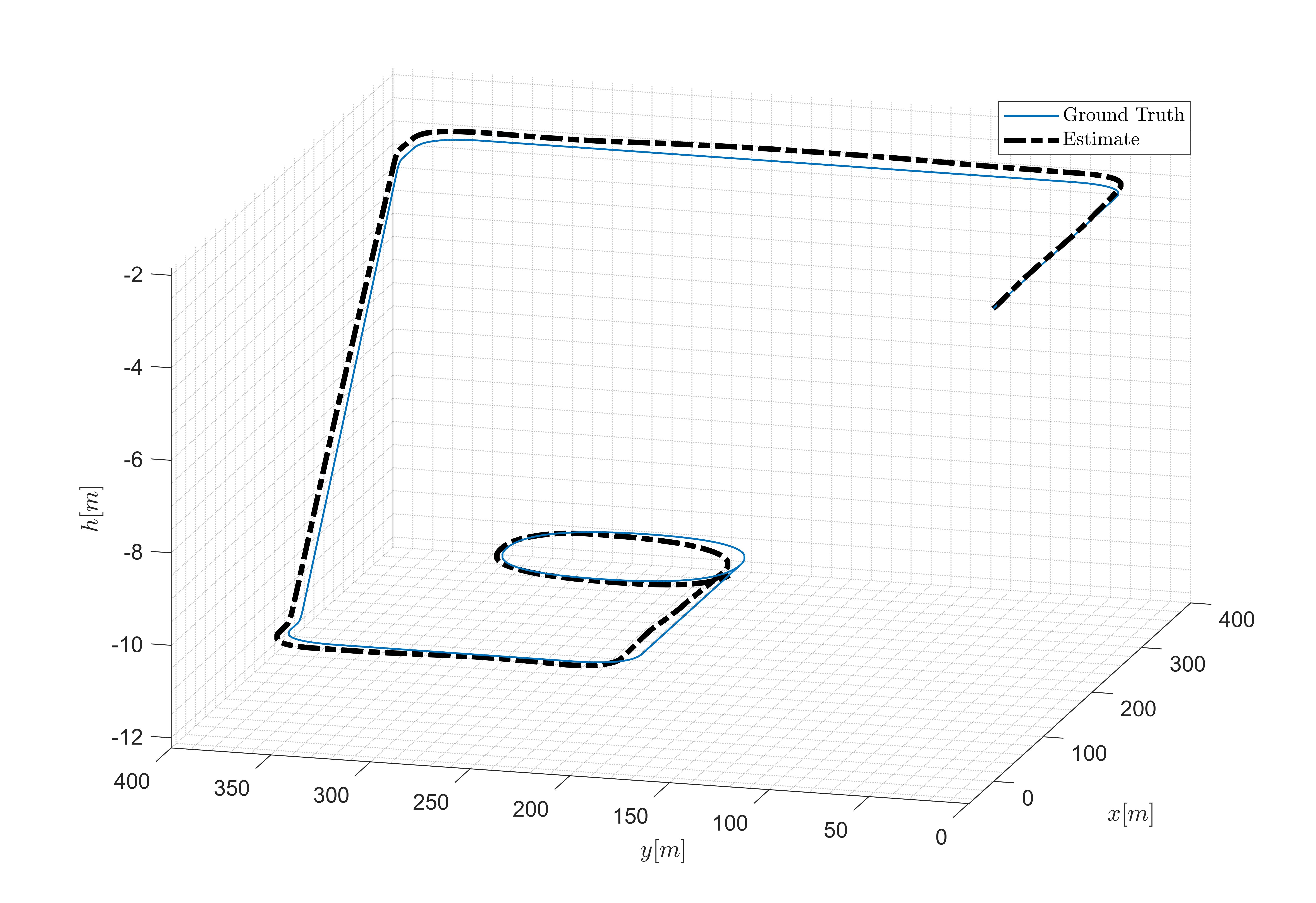}}
\caption{AUV trajectory using the velocity aided navigation filter model vs. the ground truth trajectory.}
\label{Fig31b}
\end{figure}

\begin{figure}[h!]
\centering
    {\includegraphics[width=0.4\textwidth]{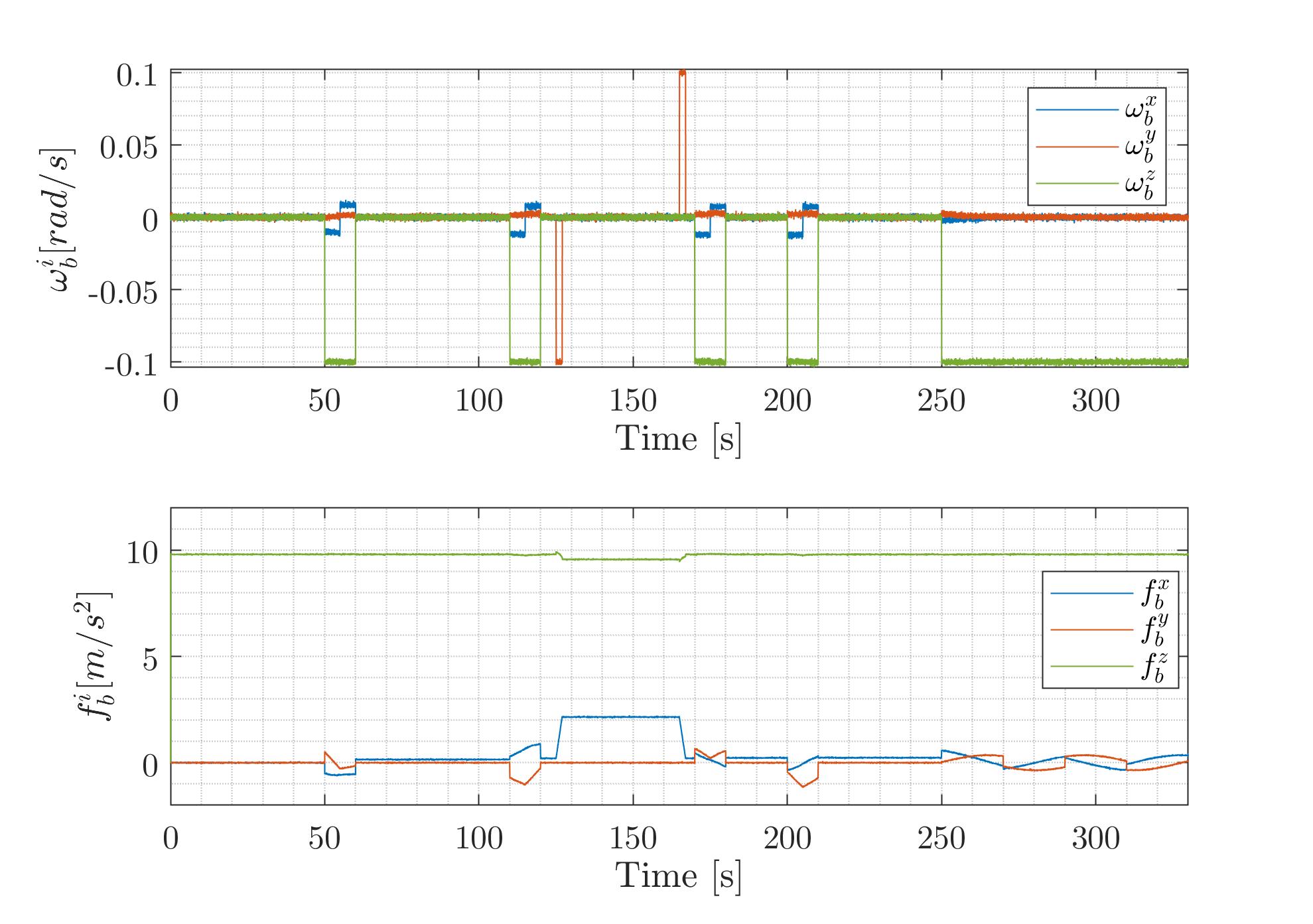}}
\caption{IMU readings of the AUV simulation.}
\label{Fig32b}
\end{figure}

All INS/DVL simulation parameters, including the DVL and IMU, are provided in Table I. 
\begin{table}[ht!]
\label{table2}
\caption {INS/DVL navigation filter parameters.} \label{tab:title} 
\setlength{\tabcolsep}{3pt}
\begin{center}
\begin{tabular}{|p{90pt}|p{60pt}|p{75pt}|}\hline
Description & Symbol & Value \\
\hline
DVL noise (var)          & $R_{11},R_{22},R_{33}$      &$0.01{[m/s]^2}$   \\ 
DVL step size &$\Delta \tau$ & $1 [s]$   \\ 
Accelerometer noise (var) & $Q_{11},Q_{22},Q_{33}$   & $0.01^2[m/s^2]^2$      \\ Gyroscope noise (var) & $Q_{44},Q_{55},Q_{66}$   & $0.001^2[rad/s]^2$ \\ Simulation duration   & $T$   &$330 [s]$    \\
Initial velocity   & ${\bf{v}}^n_0$   &$[1, 0, 0]^T [m/s]$    \\
Initial position   & ${\bf{p}}^n_0$   & $[32^0, 34^0,-5[m]]^T $    \\
\hline
\end{tabular}
\end{center}
\end{table}
To evaluate the proposed approach a typical AUV trajectory, illustrated in Figure \ref{Fig31b} was generated. An example of the inertial sensor simulated readings as a function of time are presented in (Fig. \ref{Fig32b}). 
\\
Two es-EKF with constant process noise and two model-based adaptive es-EKF were implemented and compared to the proposed approach. All cases of the model-based es-EKF, including two cases with a constant process noise covariance matrix, were examined by performing 100 Monte Carlo (MC) simulations. Results in terms of SRMSE and MAE are presented in Table II for the model and hybrid approaches.

In the first line, the process noise covariance matrix was tuned according to the true IMU covariance values. It obtained SRMSE of 1.082[m/s] and SMAE of 0.961[m/s]. Then, we examined a case where the true noise covariance matrix values are multiplied by 20. As expected, this erroneous model-based approach obtained worse results than the previous one with SRMSE of 1.142 [m/s] and SMAE of 0.992[m/s]. Applying the innovation-based approach (rows three and four), leads to an improvement of 5.5\% in the SMAE. Moving to hybrid approaches, our DNN model obtained an SRMSE of 1.055[m/s] and SMAE of 0.928[m/s], an improvement of 2.5\% and 3.4\%, respectively, over the first (constant) case. Our HCF-Ensemble hybrid model approach achieved an SRMSE of 0.98[m/s] and SMAE of 0.866[m/s], improvement of 10.4\% and 9.8\%, respectively, over the first (constant) case. It also improved the hybrid DNN approach by more than $9\%$.
\begin{table}[h!]
\caption {Experiment results for several model and learning approaches} \label{tab:title}
\begin{center}
\begin{tabular}{ |c|c|c| } 
\hline
$\bf Approach$ &${\bf  SRMSE} [m/s]$ &${\bf  SMAE} [m/s]$\\
\hline
Constant $q_f^i=0.01$ $q_{\omega}^i=0.001$  & 1.082  &0.961   \\ 
\hline
Constant $q_f^i=0.2$ $q_{\omega}^i=0.02$ &  1.142    & 0.992 \\ 
\hline
Adaptive $\xi=1$  & 1.127 &0.983  \\ 
\hline
Adaptive $\xi=5$  & 1.085 &0.908  \\ 
\hline
DNN & 1.055    & 0.928 \\ 
\hline
HCF-Ensemble & \bf{0.980}   & \bf{0.866} \\ 
\hline
\end{tabular}
\end{center}
\end{table}

\section{Conclusions} \label{sec:Conc}
The proper choice for the process noise covariance matrix is critical for accurate INS/DVL fusion. In this work, a hybrid learning model was suggested in an adaptive filter framework. To that end, a novel handcrafted features-based ensemble bagged trees model is suggested. The model learns, using a two-stage feature approach, and online tunes the system noise covariance. This learning approach was then combined with the model-based es-EKF, resulting in a hybrid adaptive navigation filter.

The DNN models and our ensemble bagged trees model were trained on a simulated database consisting of four different baseline trajectories. To validate the performance of the proposed approach, it was compared to several model-based approaches representing both constant and adaptive process noise covariance selection on an unseen AUV simulation dataset. 

Results show that our handcrafted feature ensemble model obtained an SRMSE of 0.98[m/s], an improvement of about 10\% over using a constant noise covariance matrix. This improvement also demonstrates the robustness and generalization properties of the proposed ML model. Another advantage of using an ML approach instead of a deep-learning one is a reduction of the required time to train the learning model. 

Although demonstrated for AUV INS/DVL fusion, the proposed approach can be elaborated for any external sensor aiding  to the INS and for any type of platform.

\bibliographystyle{IEEEtran}
\bibliography{IEEEfull}

\end{document}